\documentclass{article}
\usepackage{spconf,amsmath,amssymb,graphicx,cite,booktabs,hyperref}
\usepackage{algorithm}
\usepackage{algorithmic}
\usepackage{color}

\title{Plug-and-Play Diffusion Models for Image Compressive Sensing with Data Consistency Projection}

\name{Xiaodong Wang\textsuperscript{1, 2, $\dagger$}, Ping Wang\textsuperscript{1, 2, $\dagger$}, Zhangyuan Li\textsuperscript{1, 2}, Xin Yuan\textsuperscript{2,$*$}}
\address{\textsuperscript{1}Zhejiang University, Hangzhou, China \\ \textsuperscript{2,*}Westlake University, Hangzhou, China \\
\textsuperscript{$\dagger$}Equal contribution. \textsuperscript{*}Corresponding author: \texttt{xyuan@westlake.edu.cn}\\
\thanks{This work was supported by National Key R\&D Program of China (2024YFF0505603), the National Natural Science Foundation of China (grant number 62271414), Zhejiang Provincial Distinguished Young Scientist Foundation (grant number LR23F010001), Zhejiang `Pioneer' and `Leading Goose' R\&D Program (grant number 2024SDXHDX0006, 2024C03182), the Key Project of Westlake Institute for Optoelectronics (grant number 2023GD007), the 2023 International Sci-tech Cooperation Projects under the purview of the `Innovation Yongjiang 2035' Key R\&D Program (grant number 2024Z126). }}

\begin{document}
\maketitle

\begin{abstract}
We explore the connection between Plug-and-Play (PnP) methods and Denoising Diffusion Implicit Models (DDIM) for solving ill-posed inverse problems, with a focus on single-pixel imaging. We begin by identifying key distinctions between PnP and diffusion models—particularly in their denoising mechanisms and sampling procedures. By decoupling the diffusion process into three interpretable stages: denoising, data consistency enforcement, and sampling, we provide a unified framework that integrates learned priors with physical forward models in a principled manner. Building upon this insight, we propose a hybrid data-consistency module that linearly combines multiple PnP-style fidelity terms. This hybrid correction is applied directly to the denoised estimate, improving measurement consistency without disrupting the diffusion sampling trajectory. Experimental results on single-pixel imaging tasks demonstrate that our method achieves better reconstruction quality.

\end{abstract}


\section{Introduction}
Inverse problems are central to computational imaging, where the goal is to recover a latent signal $\mathbf{x}$ from corrupted measurements $\mathbf{y} = \mathbf{H(x)} + \mathbf{n}$. Such problems arise in super-resolution, deblurring, medical and computational imaging, etc., and are typically ill-posed – many possible $\mathbf{x}$ can explain the same $\mathbf{y}$. To resolve this ambiguity, prior knowledge or regularization on $\mathbf{x}$ is required. Traditional approaches formulate a regularized inverse problem (e.g. maximum a posteriori estimation) to balance data fidelity and prior plausibility.

Diffusion models (DMs) ~\cite{ho2020denoising,song2020denoising,chung2022diffusion,kawar2022denoising,wang2022zero,zheng2025inversebench,daras2024survey} have recently emerged as a powerful class of deep generative priors for images. A DM learns the distribution of clean images and generates realistic samples by iteratively denoising pure noise, outperforming GANs in fidelity and diversity. Score-based diffusion models ~\cite{song2020score} estimate the gradient of the log-density (score) of images and can draw high-quality samples via stochastic differential equations. Given their success in image synthesis, diffusion models have been explored for solving imaging inverse problems by treating the DM as a prior on $\mathbf{x}$. In parallel, plug-and-play (PnP) methods~\cite{kamilov2023plug,yuan2020plug} have become a popular optimization framework for inverse problems. PnP replaces the explicit prior in an optimization algorithm (e.g. ADMM or ISTA \cite{kamilov2023plug}) with a plugged-in image denoiser, allowing one to “play” with advanced denoisers as implicit priors. PnP schemes using DnCNN~\cite{zhang2017beyond}, FFDNet~\cite{zhang2018ffdnet} and other denoisers have achieved excellent results in super-resolution, deblurring, MRI reconstruction by iteratively enforcing measurement consistency and denoising.

Recently, researchers have begun to bridge diffusion models and PnP methods. The goal is to leverage the generative power of diffusion priors while maintaining the data-consistency flexibility of iterative PnP optimization. Early works on diffusion-based inverse problems fall into two main strategies: (1) Guidance-based diffusion, which adds a data likelihood gradient term to the diffusion sampling dynamics~\cite{chung2022diffusion,song2023pseudoinverse,wang2022zero}, and (2) Splitting-based PnP~\cite{zhu2023denoising}, which alternates between diffusion model sampling and explicit projection to the measurements. These hybrid approaches have shown promise, but also face challenges. Notably, few work has identify the connection and difference. 

To address these limitations, we propose a tighter fusion of PnP optimization and diffusion models. Rather than simply alternating or guiding diffusion with heuristic terms, we integrate a PnP-style optimization within the diffusion sampling process in a principled way. Our key idea is to enforce data fidelity constraints using plug-and-play updates at critical points in the diffusion chain, notably ensuring the final output is exactly consistent with observations. By doing so, we combine the best of both worlds: the flexibility and theoretical grounding of PnP iterative solvers and the expressive learned prior of diffusion models.

The main contributions of this work are summarized as follows:

(1) Unified PnP-Diffusion Framework: We decomposed DDIM sampling as a combination of denoising, data-consistency and sampling. By replacing the data-consistency updating in diffusion with PnP optimization steps, we build the connection between PnP and diffusion, with a special emphasis on enforcing data consistency.

(2) Hybrid Data-Consistency Enforcement in Diffusion: We introduce an algorithm that performs hybrid plug-and-play updates during diffusion sampling. This ensures that the recovered image not only is likely under the learned image prior, but also reproduces the observed measurements (addressing both manifold and measurement fidelity).

\section{Related Work}
\subsection{Plug-and-Play Methods}
Plug-and-Play (PnP) methods solve imaging inverse problems by decoupling data fidelity and prior modeling into separate iterative steps. The classical inverse problem is posed as:
\begin{equation}
\hat{\mathbf{x}} = \arg\min_{\mathbf{x} \in \mathbb{R}^n} f(\mathbf{x}) + \lambda h(\mathbf{x}),
\label{eq:pnp_objective}
\end{equation}
where $f(\mathbf{x})$ and $h(\mathbf{x})$ denote the data-fidelity term and prior-term respectively. Given a noisy observation $\mathbf{y} \in \mathbb{R}^m$ and a forward operator $\mathbf{H} \in \mathbb{R}^{m \times n}$, the data-fidelity term becomes $f(\mathbf{x})=\frac{1}{2} \| \mathbf{y} - \mathbf{H} \mathbf{x} \|_2^2$ for linear inverse problem. Proximal algorithms are often used for solving problems of the form in Eq.\ref{eq:pnp_objective} when $h(\mathbf{x})$ are nonsmooth by leveraging a mathematical concept known as the proximal operator:
\begin{equation}
\text{Prox}_{\gamma h}(\mathbf{z}) := \arg\min_{\mathbf{x} \in \mathbb{R}^n} \left\{ \frac{1}{2} \|\mathbf{x} - \mathbf{z}\|_2^2 + \gamma h(\mathbf{x}) \right\} \,,
\label{eq:2}
\end{equation}
where $\gamma \geq 0$ is an adjustable penalty parameter. PnP priors is based on the observation that the proximal operator in Eq. ~\eqref{eq:2} can be interpreted as a MAP denoiser for AWGN. All PnP frameworks can be roughly simplified as the combination of a proximal mapping on the measurement consistency $f$ and denoising term exploiting the data-prior, which is 
\begin{subequations}
\begin{align}
\mathbf{x}_k &\leftarrow \text{Prox}_{\lambda f} (\mathbf{z}_{k-1}), \label{eq:3a} \\
\mathbf{z}_k &\leftarrow \mathbf{D}_\sigma (\mathbf{x}_k),
\label{eq:3b}
\end{align}
\end{subequations}
The key idea is to iteratively alternate between \textit{data-fidelity} that enforces agreement with the measurements $\mathbf{y}$, and \textit{denoising} that serves as a proximal surrogate for the prior.

\vspace{1ex}
\noindent\textbf{PnP-HQS.} Half-Quadratic Splitting (HQS) formulation introduces an auxiliary variable $\mathbf{v}$ to decouple data and prior terms\cite{zhu2023denoising}:
\begin{equation}
    \min_{\mathbf{x}, \mathbf{v}} \ \frac{1}{2} \|\mathbf{y} - \mathbf{H}\mathbf{x}\|_2^2 + \lambda h(\mathbf{v}) \quad \text{subject to} \quad \mathbf{x} = \mathbf{v}. \label{eq:hqs}
\end{equation}
This leads to alternating updates on $\mathbf{x}$ and $\mathbf{v}$, with the latter approximated by a denoising operator $\mathcal{D}_\sigma$ in PnP.

\vspace{1ex}
\noindent\textbf{PnP-GAP.} Generalized Alternating Projection (GAP) framework imposes a hard constraint on data consistency \cite{yuan2020plug}:
\begin{equation}
    \min_{\mathbf{x}} \ h(\mathbf{x}) \quad \text{subject to} \quad \mathbf{y} = \mathbf{H}\mathbf{x}. \label{eq:gap}
\end{equation}
This can be approached by alternating projection and denoising steps.
\vspace{1ex}
\noindent The resulting iterative updates differ in the data-consistency step:
\begin{align}
    &\text{PnP-HQS:} \quad \mathbf{x}^{k+1} = \left( \mathbf{I} + \frac{1}{\gamma} \mathbf{H}^\top \mathbf{H} \right)^{-1} \left( \mathbf{v}^k + \frac{1}{\gamma} \mathbf{H}^\top \mathbf{y} \right),
 \\
    &\text{PnP-GAP:} \quad \mathbf{x}^{k+1} \leftarrow \mathbf{x}^k + \mathbf{H}^\top(\mathbf{H}\mathbf{H}^\top)^{-1} (\mathbf{y} - \mathbf{H}\mathbf{x}^k). \label{eq:pnp-gap-step} 
\end{align}
The key distinction lies in how the data-consistency term is handled: HQS solves a relaxed least-squares objective Eq. ~\eqref{eq:hqs}, while GAP projects directly onto the data constraint Eq. ~\eqref{eq:gap}. Both methods bypass the need for explicit regularizer $h(\cdot)$ by leveraging a powerful denoiser $\mathcal{D}_\sigma$ as a plug-in proximal surrogate.

\subsection{Diffusion Models}
Diffusion-based generative models aim to transform a simple known distribution (e.g., Gaussian) into a complex data distribution by learning to reverse a noise-injection process. This process can be described in continuous time using a Stochastic Differential Equation (SDE)~\cite{song2020score}:
\begin{equation}
    d\mathbf{x}_t = f(\mathbf{x}_t, t) \, dt + g(t) \, d\mathbf{W}_t,
    \label{eq:sde-forward}
\end{equation}
where $\mathbf{W}_t$ denotes a Wiener process, $f(\cdot)$ is the drift coefficient, and $g(t)$ is the diffusion coefficient. This SDE gradually perturbs the data distribution into noise as $t \to T$. Sampling from the pure noise $\mathbf{x_T}$ to the target distribution $\mathbf{x_0}$ requires solving the reverse-time SDE:
\begin{equation}
    d\mathbf{x}_t = \left[ f(\mathbf{x}_t, t) - g(t)^2 \nabla_{\mathbf{x}} \log p_t(\mathbf{x}_t) \right] dt + g(t) \, d\overline{\mathbf{W}}_t,
    \label{eq:sde-reverse}
\end{equation}
where $\nabla_{\mathbf{x}} \log p_t(\mathbf{x}_t)$ is the score function and $\overline{\mathbf{W}}_t$ is a reverse-time Wiener process.

\noindent\textbf{From SDE to DDIM.} The Denoising Diffusion Implicit Model (DDIM) \cite{song2020denoising} arises by discretizing Eq.~\ref{eq:sde-reverse} under the variance-preserving (VP) SDE setup, where $f(\mathbf{x}_t, t) = -\mathbf{x}_t$ and $g(t) = \sqrt{2}$. The VP SDE admits a closed-form forward trajectory:
\begin{equation}
    \mathbf{x}_t = \sqrt{\alpha_t} \mathbf{x}_0 + \sqrt{1 - \alpha_t} \, \mathbf{z}, \quad \mathbf{z} \sim \mathcal{N}(0, \mathbf{I}),
\end{equation}
and leads to the following efficient reverse-time update, known as DDIM sampling:
\begin{align}
\mathbf{x}_{t-1} \ & =
\sqrt{\alpha_{t-1}} 
\left(
    \frac{
        \mathbf{x}_t + (1 - \alpha_t) \nabla_{\mathbf{x}_t} \log p_t(\mathbf{x}_t)
    }{
        \sqrt{\alpha_t}
    }
\right) \nonumber \\
& + \sqrt{1 - \alpha_{t-1} - \sigma_t^2} 
\left(
    - \sqrt{1 - \alpha_t} \nabla_{\mathbf{x}_t} \log p_t(\mathbf{x}_t)
\right),
\label{eq:ddim_two_line}
\end{align}
where $\sigma_t$ controls the level of stochasticity. When $\sigma_t = 0$, this results in a purely deterministic sampling path. We can decouple this sampling as two steps: denoising and sampling step, $\mathbf{x}_{0|t}= \frac{\mathbf{x}_{t}+ (1-\alpha_{t})\,\nabla_{\mathbf{x}_{t}}\log p_{t}(\mathbf{x}_{t})}{
\sqrt{\alpha_{t}}}$ represents a denoised prediction of the clean image at time $t$, derived via Tweedie's formula. The second step is to injecting noise to the state at $\mathbf{x_{t-1}}$. The score function $\nabla_{\mathbf{x}} \log p_t(\mathbf{x}_t)$ is approximated by a neural network trained to estimate the noise or score. 

\begin{figure}[t]
    \centering
    \includegraphics[width=0.5\linewidth]{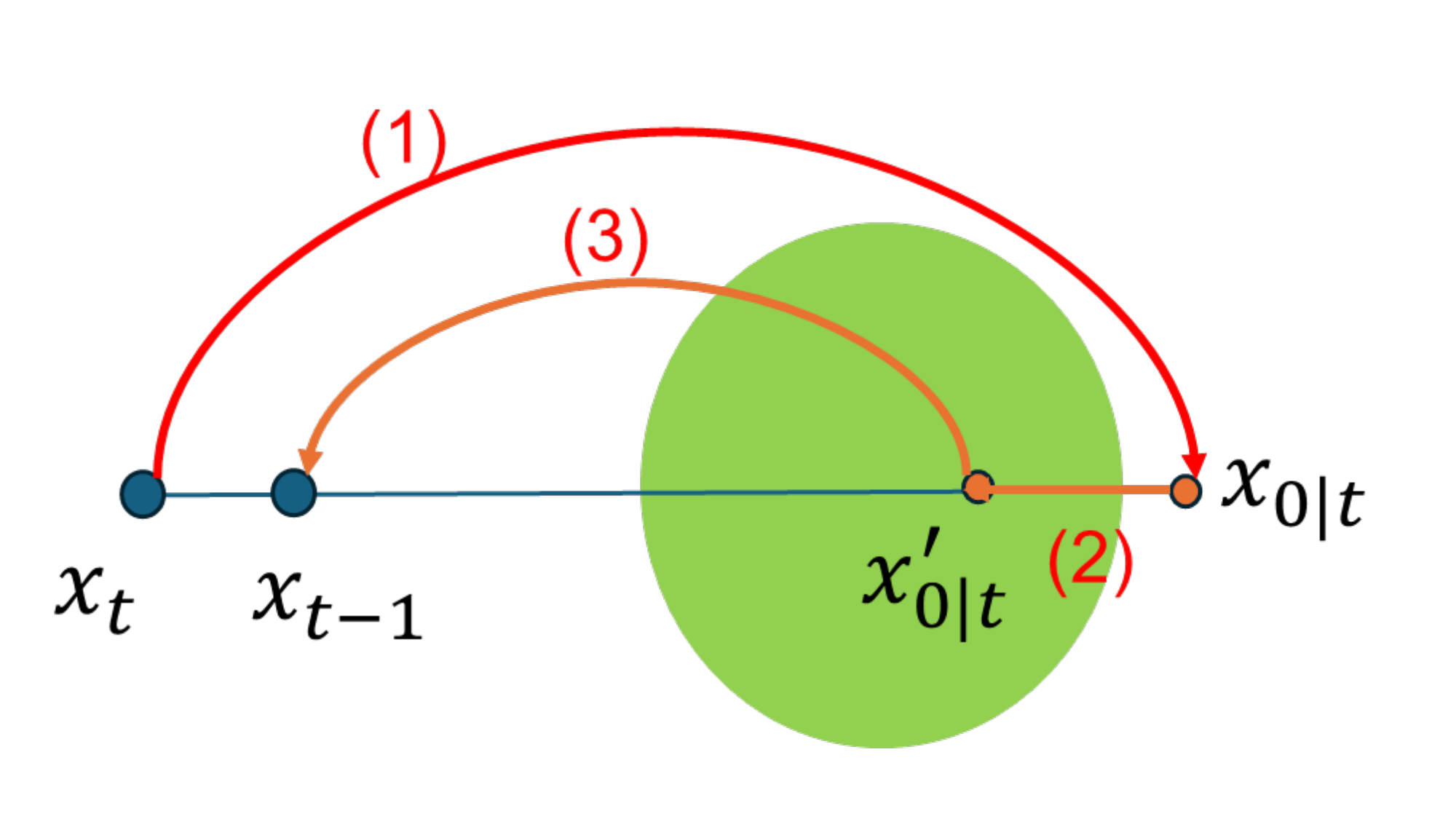}
    \vspace{-2mm}
    \caption{\small Diagram of conditional diffusion sampling step.}
    \label{fig:graph}
    \vspace{-5mm}
\end{figure}

\noindent\textbf{Conditional Sampling with DDIM.} 
In inverse problems, we are interested in sampling from the conditional distribution $p_0(\mathbf{x} \mid \mathbf{y})$. This goal can be incorporated into the diffusion framework by replacing the unconditional score $\nabla_{\mathbf{x}} \log p_t(\mathbf{x}_t)$ with the conditional score $\nabla_{\mathbf{x}} \log p_t(\mathbf{x}_t \mid \mathbf{y})$ in Eq.~\eqref{eq:ddim_two_line}. Using Bayes' rule, the conditional score can be decomposed as:
\begin{equation}
\nabla_{\mathbf{x}} \log p_t(\mathbf{x}_t \mid \mathbf{y}) 
= \nabla_{\mathbf{x}_t} \log p_t(\mathbf{x}_t) + \nabla_{\mathbf{x}_t} \log p(\mathbf{y} \mid \mathbf{x}_t),
\label{eq:conditional-score}
\end{equation}
where the first term corresponds to the learned generative score model and the second term enforces data consistency with the measurements. There are plenty of works that try to approximates $\log p(\mathbf{y} \mid \mathbf{x}_t)$, DPS~\cite{chung2022diffusion} approximates the marginal likelihood $p(\mathbf{y} \mid \mathbf{x}_t)$ as $p(\mathbf{y} \mid \mathbf{x}_t) \approx p(\mathbf{y} \mid \hat{\mathbf{x}}_0).$ IIGDM~\cite{song2023pseudoinverse} further approximates this likelihood as a Gaussian that can be easily computed. Another line of work uses variable splitting techniques to apply a proximal update to approximate the conditional posterior mean\cite{zhu2023denoising}. These methods can be implemented as a combination of a denoising term and a measurement-consistency term, which shares the same spirit with PnP principle.

\section{Method}
\subsection{Diffusion Sampling with data-consistency}
\begin{figure}[t]
    \centering
    \includegraphics[width=0.98\linewidth]{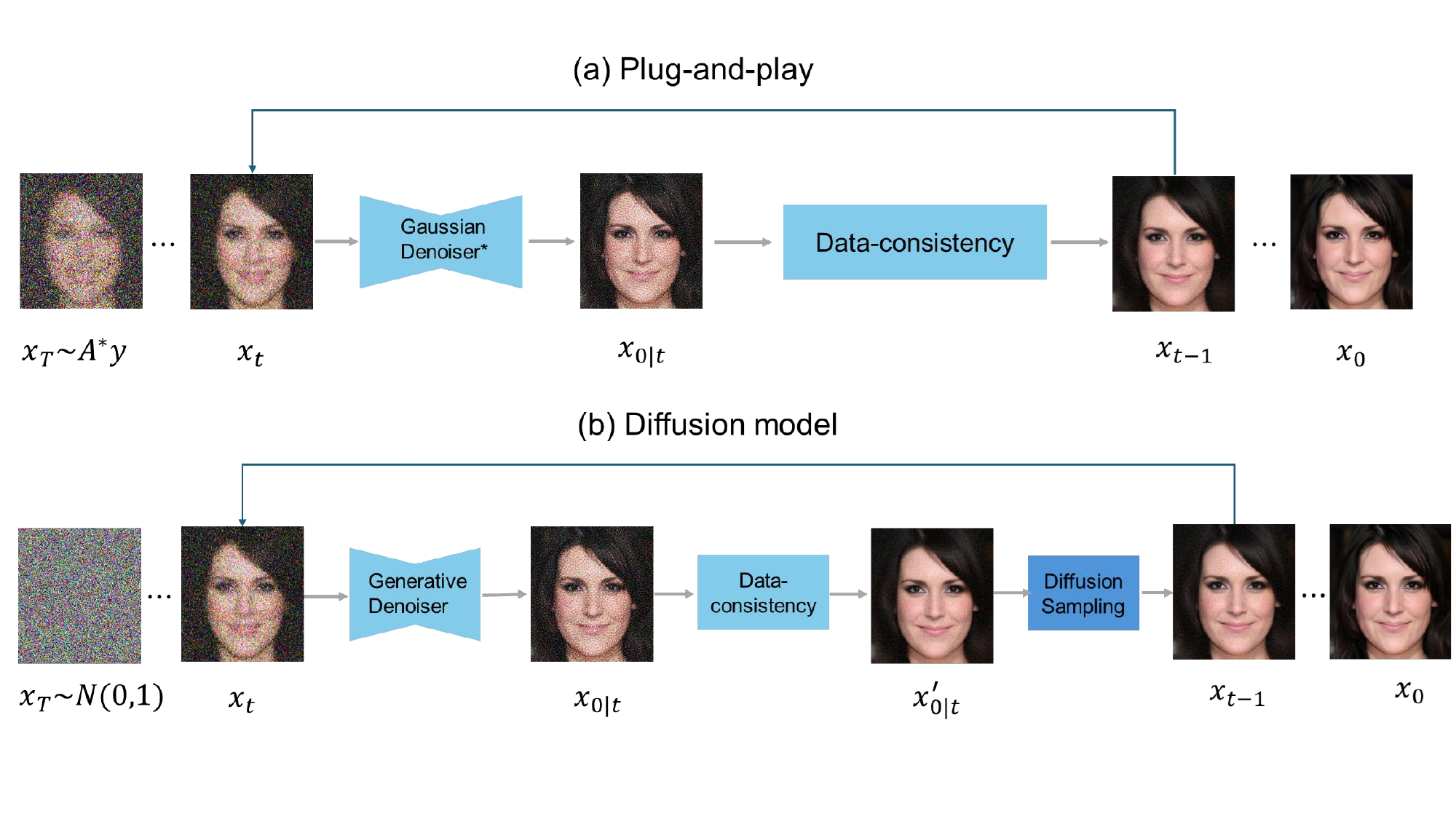}
    \vspace{-2mm}
    \caption{\small The architecture of Plug-and-play and Diffusion model for inverse problems. The difference lies in the denoiser (PnP uses a gaussian denoiser and diffusion uses generative denoiser), and the additional sampling step in diffusion model.}
    \label{fig:architecture}
    \vspace{-5mm}
\end{figure}

We explore the connection of Diffusion and PnP scheme. Both methods try to predict $\mathbf{x_{t-1}}$ from $\mathbf{x_t}$. Starting from DDIM, we decouple the conditional diffusion sampling into three steps. Substituting Eq.~\ref{eq:ddim_two_line} with Eq.\ref{eq:conditional-score} and rewrite it in a sequential manner as
\begin{subequations}
\begin{align}
\mathbf{x}_{0|t}
&= \frac{
        \mathbf{x}_{t} 
        + (1-\alpha_{t})\,\nabla_{\mathbf{x}_{t}}\log p_{t}(\mathbf{x}_{t})
     }{
        \sqrt{\alpha_{t}}
     },
\tag{13a} \label{eq:ddim-denoise3} \\[1ex]
\mathbf{x}_{0|t}'&= \mathbf{x}_{0|t}-\mu_t \nabla_{\mathbf{x_t}} \log p(\mathbf{y} \mid \mathbf{x}_t),
\tag{13b} \label{eq:ddim-denoise3} \\[1ex]
\mathbf{x}_{t-1}
&= \sqrt{\alpha_{t-1}}\, \mathbf{x}_{0|t}'
\nonumber \\[-0.5ex]
&\quad
- \sqrt{1 - \alpha_{t-1} - \sigma_t^2}\,
\left(
    \sqrt{1 - \alpha_t} \nabla_{\mathbf{x}_t} \log p_t(\mathbf{x}_t)
\right),
\tag{13c} \label{eq:ddim-sample3}
\end{align}
\end{subequations}
where $\mu_t=\sqrt{\frac{(1-\alpha_{t-1}-\sigma_t^2)(1-\alpha_t)}{\alpha_t}}$ represents the step size. As can be seen in Fig.~\ref{fig:graph}, we adapt the condition diffusion sampling as a more general form by decoupling the DDIM sampling process into three-steps:
\begin{itemize}
    \item[(1)] \textbf{Denoising step:} Estimate a clean latent image $\mathbf{x}_{0|t}$ using a pretrained denoiser or score.
    \item[(2)] \textbf{Data-consistency step:} Correct the denoised estimate using the measurement model to $\mathbf{x}_{0|t}'$ with more measurement consistency.
    \item[(3)] \textbf{Sampling step:} Generate the next sample $\mathbf{x}_{t-1}$ using the corrected latent estimate.
\end{itemize}
We can see that DDIM share the same spirit with PnP by iterating between a denoising and measurement projection term, where we use the proximal operator in PnP framework to represent the measurement-consistency term. This has a similar form as illustrated in ~\cite{zhu2023denoising} when the proximal operator is HQS operator. In this way, we formulate the explicit link between diffusion model and PnP for inverse problem through the data-consistency term. In the following, similar to ~\cite{zhu2023denoising}, we use PnP data-consistency updating to substitute data-consistency step in diffusion. The difference between diffusion model and PnP is illustrated in Fig.~\ref{fig:architecture}, where in diffusion model a scheduled denoiser and a sampling step are the main difference. To further illustrate the difference between this work and our method, we further propose a hybrid form of data-consistency updating to exploit the soft and hard constraint imposed by HQS and GAP. This exploration is verified using a single-pixel imaging system.

\subsection{Hybrid Plug-and-Play Data Consistency}

Inspired by \cite{garber2024image}, we propose to enhance measurement consistency in diffusion restoration by linearly combining PnP data-consistency terms. Specifically, we integrate the closed-form updates of PnP-GAP and PnP-HQS into a fused formulation.
\textbf{GAP Update.} The GAP update enforces strict measurement consistency via back-projection:
\begin{equation}
\mathbf{x}_{{0|t}}' = \mathbf{x}_{0|t} + \mathbf{H}^{\dagger}(\mathbf{y} - \mathbf{H}\mathbf{x}_{0|t}),
\end{equation}
where $\mathbf{H}^{\dagger} = \mathbf{H}^\top(\mathbf{H}\mathbf{H}^\top)^{-1}$ is the Moore-Penrose pseudoinverse.

\noindent\textbf{HQS Update.} The HQS update solves a regularized least-squares problem:
\begin{equation}
\mathbf{x}_{{0|t}}' = (\mathbf{H}^T\mathbf{H} + \lambda\mathbf{I})^{-1}(\mathbf{H}^T\mathbf{y} + \lambda \mathbf{x}_{0|t}),
\end{equation}
where $\lambda$ balances the regularization.

\noindent\textbf{Fused Update.} We combine the two updates:
\begin{align}
\mathbf{x}_{0|t}' &= (1 - \delta_t)(\mathbf{x}_{0|t} + \mathbf{H}^{\dagger}(\mathbf{y} - \mathbf{H}\mathbf{x}_{0|t})) \nonumber \\ &+ \delta_t(\mathbf{H}^\top\mathbf{H} + \lambda\mathbf{I})^{-1}(\mathbf{H}^\top\mathbf{y} + \lambda \mathbf{x}_{0|t}),
\label{eq:fusedprox}
\end{align}
where $\delta_t \in [0,1]$ controls the fusion between strict and soft constraints. This fusion can be interpreted as a preconditioned gradient step, improving robustness and convergence in the linear inverse problem.
\subsection{Compressive Sensing Inverse Problem}
We consider the inverse problem of single-pixel imaging (SPI)~\cite{zhang2018ista}. Instead of using a typical SPI forward model where block compressed sensing is employed to reduce the workload of computing the sensing matrix $\mathbf{H}$, we turn to \cite{wang2023saunet} to use a 2D CS model to avoid multi-block computation. Its compressed measurement is obtained via:
\begin{equation}
\mathbf{Y} = \mathbf{U}\mathbf{X}\mathbf{V}^\top,
\end{equation}
where $\mathbf{U} \in \mathbb{R}^{\sqrt{M} \times \sqrt{N}}$ and $\mathbf{V} \in \mathbb{R}^{\sqrt{M} \times \sqrt{N}}$ be two independent sensing matrices applied along the horizontal and vertical directions, respectively. This yields a separable measurement model that compresses a 2D signal simultaneously in both dimensions. By using the Kronecker product, we express the 2D measurement equation in a vectorized form:
\begin{equation}
\mathbf{y} = \mathbf{H}\mathbf{x}, \quad \text{s.t.} \quad 
\begin{cases}
\mathbf{x} = \operatorname{vec}(\mathbf{X}) \\
\mathbf{y} = \operatorname{vec}(\mathbf{Y}) \\
\mathbf{H} = \mathbf{U} \otimes \mathbf{V}
\end{cases}
\end{equation}
Here, $\operatorname{vec}(\cdot)$ denotes the vectorization operator that stacks a matrix column-wise into a vector, and $\otimes$ denotes the Kronecker product. This formulation allows us to represent 2D acquisition in a linear inverse problem form. In the CS framework, suppose the measurement matrix $\mathbf{H} = \mathbf{U} \otimes \mathbf{V}$ satisfies the orthogonality condition $\mathbf{H} \mathbf{H}^\top = \mathbf{I}$, which implies $\mathbf{U} \mathbf{U}^\top \otimes \mathbf{H} \mathbf{H}^\top = \mathbf{I}$. Under this assumption, the fused update in Eq. \ref{eq:fusedprox} can be approximated as:
\begin{align}
\mathbf{x}_{0|t}' &=  \mathbf{x}_{0|t} + \rho\mathbf{U}^{\top}(\mathbf{y}-\mathbf{U}\mathbf{x}_{0|t}\mathbf{V}^{\top})\mathbf{V}, 
\label{eq:fuseprox_final}
\end{align}
where $\rho = 1 - \frac{\lambda\delta_t}{1+\lambda}$. Eq.~\eqref{eq:fuseprox_final} can be interpreted as a projected gradient descent update enforcing data fidelity based on the separable structure of the sensing model.

\begin{algorithm}[H]
\caption{Fused Data Guidance for Diffusion Sampling}
\begin{algorithmic}[1]
\REQUIRE Observation $\mathbf{y}$, SPI forward model $\mathbf{H}$, score function, diffusion schedule $\{\sigma_t\}$, fusion weights $\{\delta_t\}$
\STATE Initialize $\mathbf{x}_T \sim \mathcal{N}(0, \mathbf{I})$
\FOR{$t = T$ to $1$}
    \STATE $\mathbf{x}_{0\mid t} \leftarrow \frac{
        \mathbf{x}_{t} 
        + (1-\alpha_{t})\,\nabla_{\mathbf{x}_{t}}\log p_{t}(\mathbf{x}_{t})
     }{
        \sqrt{\alpha_{t}}
     }$ \hfill // Denoise
    \STATE $\mathbf{g}_{\text{GAP}} \leftarrow  \mathbf{x}_{0|t} + \mathbf{H}^{\dagger}(\mathbf{y} - \mathbf{H}\mathbf{x}_{0|t})$ \hfill // GAP term
    \STATE $\mathbf{g}_{\text{HQS}} \leftarrow (\mathbf{H}^T\mathbf{H} + \lambda\mathbf{I})^{-1}(\mathbf{H}^T\mathbf{y} + \lambda \mathbf{x}_{0|t})$ \hfill // HQS term
    \STATE $\mathbf{x}_{0\mid t}' \leftarrow (1 - \delta_t)\mathbf{g}_{\text{GAP}} + \delta_t\mathbf{g}_{\text{HQS}}$\hfill// Fused term
    \STATE $\hat{\epsilon}_t \leftarrow \frac{ \left( \mathbf{x}_t - \sqrt{\alpha_t} \, \mathbf{x}_{0\mid t}' \right)}{\sqrt{1 - \alpha_t}}$
    \STATE $\boldsymbol{\epsilon}_t \sim \mathcal{N}(0, \mathbf{I}_n)$
    \STATE \small
    $\begin{aligned}[t]
    \mathbf{x}_{t-1} &= \sqrt{\bar{\alpha}_{t-1}} \, \mathbf{x}'_{0|t} \\
    &\quad + \sqrt{1 - \bar{\alpha}_{t-1}} \left( w_t \sqrt{1 - \zeta} \, \hat{\boldsymbol{\epsilon}}_t + \sqrt{\zeta} \, \boldsymbol{\epsilon}_t \right)
    \end{aligned}$
    \STATE // DDIM Sampling
    \normalsize
\ENDFOR
\RETURN $\mathbf{x}_0$
\label{alfo}
\end{algorithmic}
\end{algorithm}

\section{Experiments}

We evaluate the proposed fusion strategy on Single-Pixel Imaging, a compressed sensing system where only a small number of linear projections of the scene are measured through a binary modulation matrix $\mathbf{H}$. To thoroughly assess performance under varying levels of measurement sparsity, we simulate five compression rates: 1\%, 5\%, 10\% and 20\%.

\subsection{Datasets and Implementation Details}
\noindent\textbf{Implementation Details.} We evaluate the proposed reconstruction in an extreme compressive sensing scenario: single-pixel imaging. In SPI, a single photodiode measures the scene intensity modulated by a series of known spatial patterns, rather than using a full image sensor. We simulate this by applying $M$ sequential binary mask patterns to each test image and recording one intensity value per pattern.We consider five compression ratios (CRs): 1\%, 5\%, 10\% and 20\%. A set of 18 natural images from McMaster dataset ~\cite{zhang2011color} (resized to $256\times256$) is used as the test set. As our method is zero-shot, we use pre-trained models from ~\cite{dhariwal2021diffusion} trained on ImageNet. We use T = 100 iterations in DDIM sampling for each of the methods. The algorithm framework is illustrated in Algo 1. The choice of diffusion schedule $\{\sigma_t\}$ and fusion weights $\{\delta_t\}$ are consistent with \cite{garber2024image}.

\begin{figure}[t]
    \centering
    \includegraphics[width=0.98\linewidth]{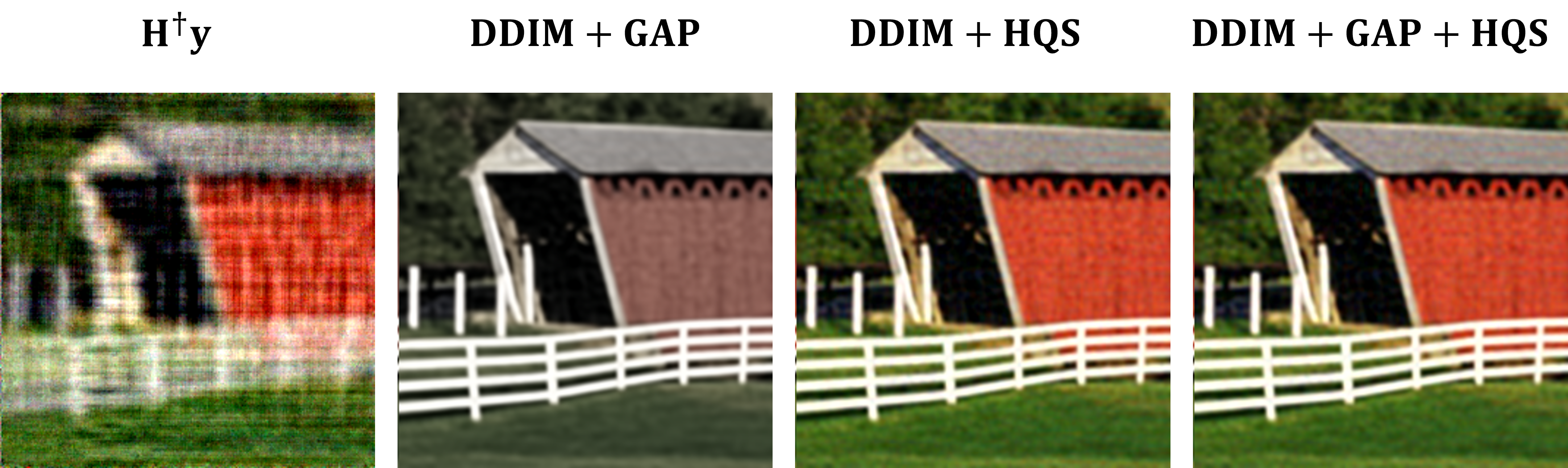}
    \vspace{-2mm}
    \caption{\small Single-pixel imaging using our hybrid diffusion-PnP method.}
    \label{fig:vis}
    \vspace{-2mm}
\end{figure}

\noindent\textbf{Comparison with other methods.} We compare our fused method against baseline zero-shot diffusion methods, including $\mathbf{H}^{\dagger}\mathbf{y}$, $\mathbf{DDIM+GAP}$, $\mathbf{DDIM+HQS}$ and a hybrid diffusion sampling $\mathbf{DDIM+GAP+HQS}$, which represents a combination of DDIM with corresponding data-consistency updating. Here, we find the data-consistency in GAP and HQS resembles the guidance-based diffusion model DDNM\cite{wang2022zero} and variable-splitting method DiffPIR\cite{zhu2023denoising} respectively. Therefore, we can think of our hybrid method as a linear combination of DDNM and DiffPIR. Evaluation metrics include Peak Signal-to-Noise Ratio (PSNR), Structural Similarity Index (SSIM), and Learned Perceptual Image Patch Similarity (LPIPS) distance. LPIPS measures the perceptual similarity between two images. PSNR and SSIM measures the faithfulness of restoration between two images.

\begin{figure}[t]
    \centering
    \includegraphics[width=0.999\linewidth]{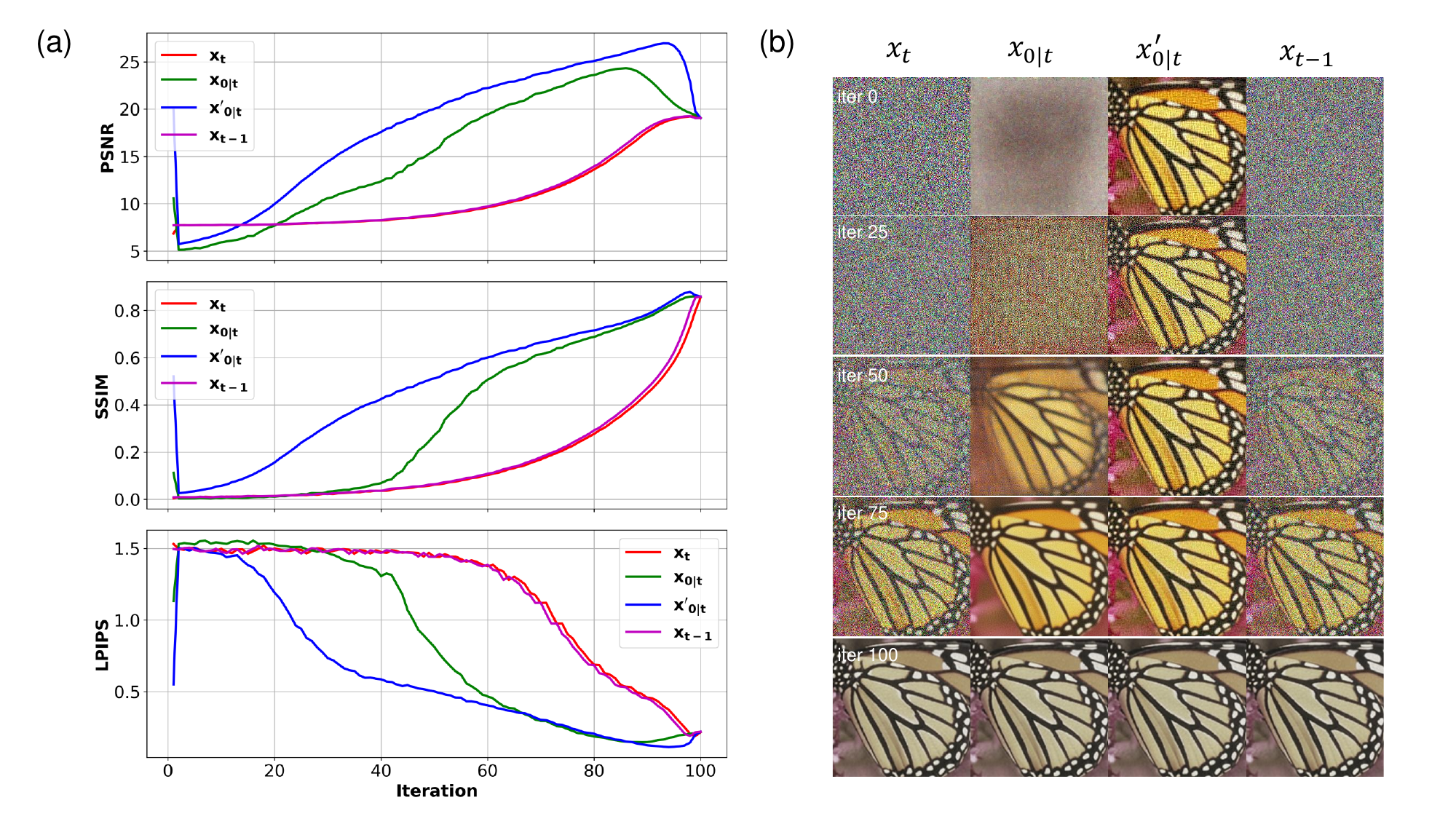}
    \vspace{-2mm}
    \caption{\small Visulization of data guidance for diffusion sampling. (a) Metrics evolving in diffusion process. (b) Visualization of diffusion sampling with data-consistency enforcement. It can be seen that $\mathbf{x}_{0\mid t}'$ enforce measurement consistent.}
    \label{fig:process}
    \vspace{-5mm}
\end{figure}

\subsection{Quantitative Experiments}
We compare our method with other diffusion model. Experimental results in Tab.~\ref{tab:algo} show that our fused data-guided diffusion outperforms baselines with compressed ratio of 5\%. We show in Fig.~\ref{fig:vis} that our hybrid combination of GAP and HQS leads to better performance compared with separate data-consistency term. It can be seen that diffusion model with GAP consistency leads to degraded color whereas our hybrid method reconstruct high-quality image with better color correction. The color distortion in GAP may arise since GAP as a hard constraint is not robust to noise distortion.

Tab. ~\ref{tab:results} summarizes the reconstruction accuracy for single-pixel compressive sensing at compression ratios ranging from 1\% to 20\%. In conclusion, our method combines physical consistency with powerful generative priors and remains effective even when very few measurements are available, making it well-suited for practical SPI deployments with stringent acquisition budgets. We also show the sampling process of intermediate steps. As shown in Fig.~\ref{fig:process}, the hybrid method tends to drop performance at the beginning and then catch up the reconstruction process later on in all three metrics. This could be further explored in the future.
\begin{table}[h]
\vspace{-5mm}
\centering
\caption{Reconstruction metrics comparison across methods.}
\vspace{2mm}
\begin{tabular}{lcccc}
\toprule
\textbf{Method} & \textbf{PSNR (dB)} & \textbf{SSIM} & \textbf{LPIPS} \\
\midrule
$\mathbf{H}^{\dagger}\mathbf{y}$ &  20.55 & 0.39 & 0.54 \\
DDIM+GAP & 24.09 & 0.62 & 0.33 \\
DDIM+HQS & 24.64 & 0.67 & 0.38 \\
DDIM+GAP+HQS & 24.76 & 0.68 & 0.37 \\
\bottomrule
\label{tab:algo}
\end{tabular}
\end{table}
\vspace{-10mm}
\begin{table}[h]
\vspace{-3mm}
\centering
\caption{Reconstruction results across different CRs.}
\vspace{2mm}
\begin{tabular}{lccc}
\toprule
CR & PSNR & SSIM & LPIPS \\
\midrule
1\% & 21.23 & 0.47 & 0.56 \\
5\% & 24.76 & 0.68 & 0.37 \\
10\% & 25.66 & 0.70 & 0.25 \\
20\% & 27.01 & 0.78 & 0.18 \\
\bottomrule
\label{tab:results}
\end{tabular}
\end{table}
\vspace{-10mm}

\section{Conclusion}

In this work, we investigated the relationship between Plug-and-Play and diffusion model for ill-posed inverse problems, with a particular focus on single-pixel imaging. By analyzing the structural differences in denoising strategies and sampling dynamics, we decoupled the diffusion process into three modular components: denoising, data consistency, and sampling. This perspective allowed us to construct a unified framework that systematically integrates learned diffusion priors with physical measurement models. To further enhance measurement fidelity, we introduced a hybrid data-consistency module that combines multiple PnP-style updates in a linear fusion scheme. This hybrid correction is directly applied to the clean denoised estimate $\mathbf{x}_{0|t}$, thereby improving alignment with observations while preserving the stochastic nature of diffusion-based sampling. Experiments on single-pixel imaging tasks validate the effectiveness of our approach. The proposed method consistently outperforms DDNM and DiffPIR baselines. Our results demonstrate the potential of fusing principled physical constraints with modern generative models for robust image recovery. 

\bibliographystyle{IEEEbib}
\bibliography{refs}

\begin{thebibliography}{10}

\bibitem{ho2020denoising}
Jonathan Ho, Ajay Jain, and Pieter Abbeel,
\newblock ``Denoising diffusion probabilistic models,''
\newblock {\em Advances in neural information processing systems}, vol. 33, pp. 6840--6851, 2020.

\bibitem{song2020denoising}
Jiaming Song, Chenlin Meng, and Stefano Ermon,
\newblock ``Denoising diffusion implicit models,''
\newblock {\em arXiv preprint arXiv:2010.02502}, 2020.

\bibitem{chung2022diffusion}
Hyungjin Chung, Jeongsol Kim, Michael~T Mccann, Marc~L Klasky, and Jong~Chul Ye,
\newblock ``Diffusion posterior sampling for general noisy inverse problems,''
\newblock {\em arXiv preprint arXiv:2209.14687}, 2022.

\bibitem{kawar2022denoising}
Bahjat Kawar, Michael Elad, Stefano Ermon, and Jiaming Song,
\newblock ``Denoising diffusion restoration models,''
\newblock {\em Advances in Neural Information Processing Systems}, vol. 35, pp. 23593--23606, 2022.

\bibitem{wang2022zero}
Yinhuai Wang, Jiwen Yu, and Jian Zhang,
\newblock ``Zero-shot image restoration using denoising diffusion null-space model,''
\newblock {\em arXiv preprint arXiv:2212.00490}, 2022.

\bibitem{zheng2025inversebench}
Hongkai Zheng, Wenda Chu, Bingliang Zhang, Zihui Wu, Austin Wang, Berthy~T Feng, Caifeng Zou, Yu~Sun, Nikola Kovachki, Zachary~E Ross, et~al.,
\newblock ``Inversebench: Benchmarking plug-and-play diffusion priors for inverse problems in physical sciences,''
\newblock {\em arXiv preprint arXiv:2503.11043}, 2025.

\bibitem{daras2024survey}
Giannis Daras, Hyungjin Chung, Chieh-Hsin Lai, Yuki Mitsufuji, Jong~Chul Ye, Peyman Milanfar, Alexandros~G Dimakis, and Mauricio Delbracio,
\newblock ``A survey on diffusion models for inverse problems,''
\newblock {\em arXiv preprint arXiv:2410.00083}, 2024.

\bibitem{song2020score}
Yang Song, Jascha Sohl-Dickstein, Diederik~P Kingma, Abhishek Kumar, Stefano Ermon, and Ben Poole,
\newblock ``Score-based generative modeling through stochastic differential equations,''
\newblock {\em arXiv preprint arXiv:2011.13456}, 2020.

\bibitem{kamilov2023plug}
Ulugbek~S Kamilov, Charles~A Bouman, Gregery~T Buzzard, and Brendt Wohlberg,
\newblock ``Plug-and-play methods for integrating physical and learned models in computational imaging: Theory, algorithms, and applications,''
\newblock {\em IEEE Signal Processing Magazine}, vol. 40, no. 1, pp. 85--97, 2023.

\bibitem{yuan2020plug}
Xin Yuan, Yang Liu, Jinli Suo, and Qionghai Dai,
\newblock ``Plug-and-play algorithms for large-scale snapshot compressive imaging,''
\newblock in {\em Proceedings of the IEEE/CVF Conference on Computer Vision and Pattern Recognition}, 2020, pp. 1447--1457.

\bibitem{zhang2017beyond}
Kai Zhang, Wangmeng Zuo, Yunjin Chen, Deyu Meng, and Lei Zhang,
\newblock ``Beyond a gaussian denoiser: Residual learning of deep cnn for image denoising,''
\newblock {\em IEEE transactions on image processing}, vol. 26, no. 7, pp. 3142--3155, 2017.

\bibitem{zhang2018ffdnet}
Kai Zhang, Wangmeng Zuo, and Lei Zhang,
\newblock ``Ffdnet: Toward a fast and flexible solution for cnn-based image denoising,''
\newblock {\em IEEE Transactions on Image Processing}, vol. 27, no. 9, pp. 4608--4622, 2018.

\bibitem{song2023pseudoinverse}
Jiaming Song, Arash Vahdat, Morteza Mardani, and Jan Kautz,
\newblock ``Pseudoinverse-guided diffusion models for inverse problems,''
\newblock in {\em International Conference on Learning Representations}, 2023.

\bibitem{zhu2023denoising}
Yuanzhi Zhu, Kai Zhang, Jingyun Liang, Jiezhang Cao, Bihan Wen, Radu Timofte, and Luc Van~Gool,
\newblock ``Denoising diffusion models for plug-and-play image restoration,''
\newblock in {\em Proceedings of the IEEE/CVF Conference on Computer Vision and Pattern Recognition}, 2023, pp. 1219--1229.

\bibitem{garber2024image}
Tomer Garber and Tom Tirer,
\newblock ``Image restoration by denoising diffusion models with iteratively preconditioned guidance,''
\newblock in {\em Proceedings of the IEEE/CVF Conference on Computer Vision and Pattern Recognition}, 2024, pp. 25245--25254.

\bibitem{zhang2018ista}
Jian Zhang and Bernard Ghanem,
\newblock ``Ista-net: Interpretable optimization-inspired deep network for image compressive sensing,''
\newblock in {\em Proceedings of the IEEE conference on computer vision and pattern recognition}, 2018, pp. 1828--1837.

\bibitem{wang2023saunet}
Ping Wang and Xin Yuan,
\newblock ``Saunet: Spatial-attention unfolding network for image compressive sensing,''
\newblock in {\em Proceedings of the 31st ACM International Conference on Multimedia}, 2023, pp. 5099--5108.

\bibitem{zhang2011color}
Lei Zhang, Xiaolin Wu, Antoni Buades, and Xin Li,
\newblock ``Color demosaicking by local directional interpolation and nonlocal adaptive thresholding,''
\newblock {\em Journal of Electronic imaging}, vol. 20, no. 2, pp. 023016--023016, 2011.

\bibitem{dhariwal2021diffusion}
Prafulla Dhariwal and Alexander Nichol,
\newblock ``Diffusion models beat gans on image synthesis,''
\newblock {\em Advances in neural information processing systems}, vol. 34, pp. 8780--8794, 2021.

\end{thebibliography}

\end{document}